\documentclass[11pt,letterpaper]{article}
\usepackage[T1]{fontenc}
\usepackage[utf8]{inputenc}
\usepackage{lmodern}
\usepackage[margin=0.8in]{geometry}
\usepackage{graphicx,booktabs,longtable,array,tabularx,float}
\usepackage{amsmath,microtype}
\usepackage{xurl}
\usepackage[hidelinks,unicode]{hyperref}
\usepackage[font=small,labelfont=bf]{caption}
\usepackage[section]{placeins}
\title{Decision models for UAVBench and the viability of edge deployment}
\author{Matthew Michelson \and Goran Muric \and Adrien Bibal \and Steven Minton}
\date{\small InferLink Corporation, Marina del Rey, CA}
\begin{document}
\maketitle
\begin{abstract}
Small language models are often proposed for reasoning onboard a drone. However, a good benchmark score does not tell us whether a model will actually run, and run usefully, on an edge computer. In this paper we study three related questions using the UAV decision benchmark, UAVBench. First, are small structured decision models good candidates for these tasks? Second, can an ordinary local machine (a Mac) screen candidate models before we test them on physical devices? Third, what software and resource limits govern deployment? Our main Mac evaluation covers 24 model configurations on the same 2,000 validated multiple-choice questions from UAVBench. These include 20 size-and-precision configurations of 11 generative models, plus four open-weight structured decision models that received no additional UAVBench-specific training. A structured decision model, H2O Lightning 4B, achieved the highest accuracy we observed, scoring 77.65\% on the Mac. This shows that these models are well suited to this kind of task. However, when we tried to deploy the decision models, we ran into conversion and runtime limits on the edge devices we tested, which we describe in detail. To test local screening, we ran matched evaluations of two generative models on both the Mac and a Qualcomm RB3 board. The accuracies did not match exactly, but the ranking did. LFM2 2.6B ranked above Qwen2.5 1.5B on both platforms. This supports using a local machine to shortlist candidates. That said, individual answers still differed between platforms, so testing on the actual hardware remained necessary.

\end{abstract}

\section{Introduction}\label{sec:1}
A drone may need to choose among a few mission actions based on a description of its current state. For instance, it might need to decide whether to continue a task, return to base, or request assistance. To be clear, in this use case the decision model runs on a different computer than the flight controller. It does not handle low-level activities such as stabilization or acceleration. Rather, it decides which high-level commands are executed. The practical question is which small models can make these choices accurately enough, quickly enough, and on a computer small enough for the drone to carry.

As a concrete example, consider the UAVBench question in Figure 1, taken from the benchmark's validated release [1,2]. A fixed-wing drone is flying a search-and-rescue grid near an airport, in a crosswind that grows stronger with altitude, with degraded satellite positioning (GNSS) and a 30\% battery reserve. The question asks which of seven actions keeps the search both efficient and safe. Some options break the rules outright. For instance, climbing to 180 m exceeds the 150 m ceiling, and descending to 25 m goes below the 30 m floor. Others are plausible but wrong, such as returning immediately. The correct answer, D, is to drop to 60 m and adjust the grid spacing. We use this question as a running example.

\begin{figure}[H]\centering\includegraphics[width=\linewidth]{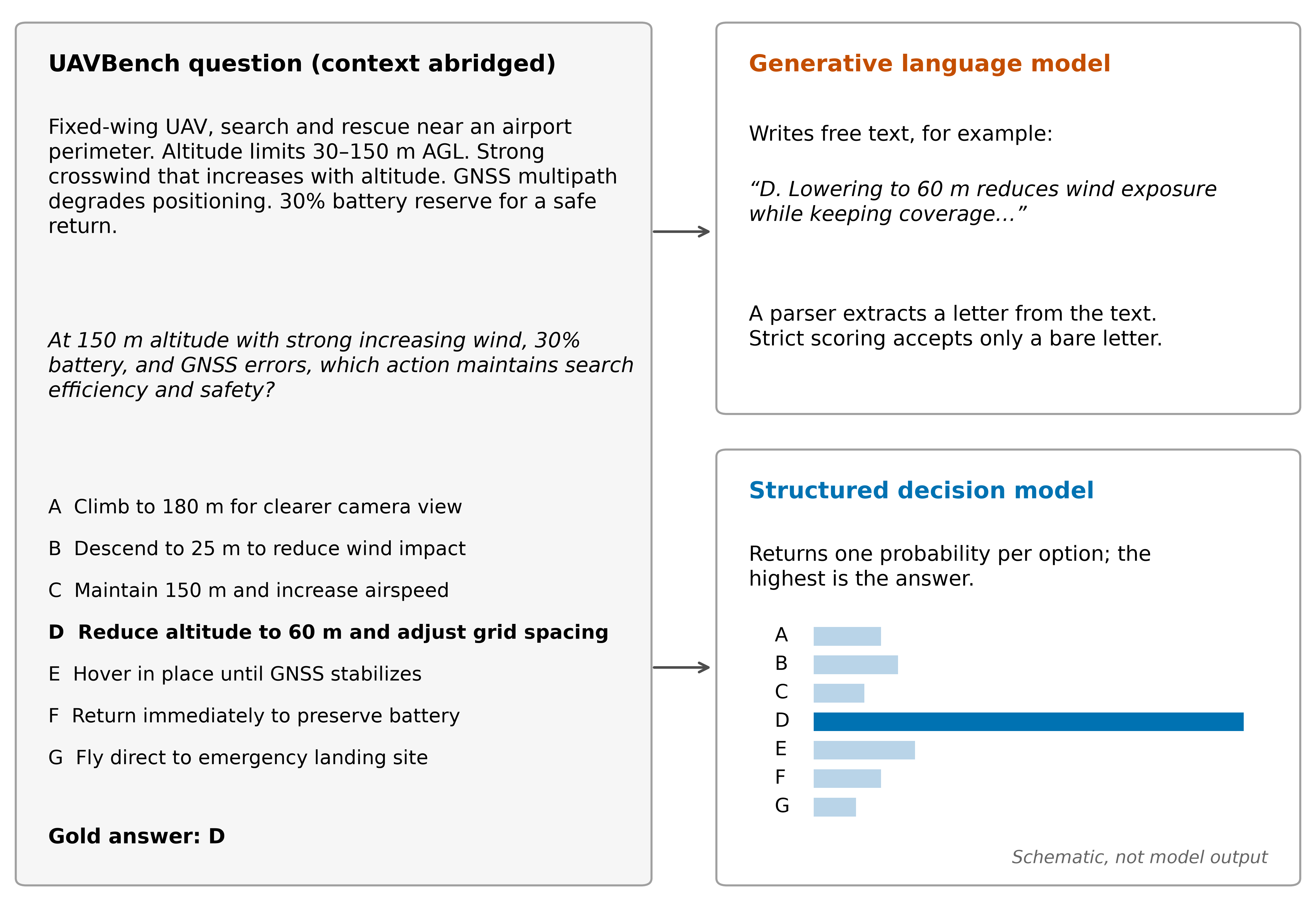}\caption{A UAVBench question (style 10, hybrid integrated reasoning) and the two ways a model can answer it. A generative model writes text, and a parser extracts a letter. A structured decision model returns a probability for each option. The generated text and the bar heights are schematic, not outputs of a tested model.}\label{fig:1}\end{figure}
Figure 1 also shows the two ways a model can answer. A generative language model writes text, and a parser then has to find the chosen letter in it. A structured decision model (sometimes called a typed decision model) skips this step. Given a state, a question and the named options, it returns a probability for each option, and the most probable option is the answer. TypeSafe AI offers this interface commercially through its Jev model, which it calls a System One model [3]. More recently, open-weight releases have made this kind of model easy to test on a local machine [4--7]. A finite-choice benchmark like UAVBench is a natural place to test them, because the required output is already a choice among options.

In this paper, we ask three questions. First, are small structured decision models good candidates for UAV mission decisions? Second, testing on real edge hardware is slow and costly, so can an ordinary desktop machine (in our case, a Mac) rank candidates first? Third, what gets in the way when we try to deploy the best candidates on a physical robotics board? To answer them, we evaluate 24 model configurations on a Mac, repeat matched experiments on a Qualcomm RB3 Gen 2 development board to test if the rank-order of model accuracy is preserved, and third attempt to port each decision model to the RB3.

Put simply, choosing a model and deploying it are connected questions, and our main contribution is to link them. In summary, this paper makes three contributions:

We show that four open-weight structured decision models perform well on UAVBench without any task-specific training. Three of the four score higher than every one of the 20 generative configurations we tested, and the best scores 77.65\%.

We show that a local machine gives a useful first ranking before testing on edge hardware. However, matched experiments also show that similar overall accuracy can hide reproducible changes in individual answers, so hardware validation is still needed.

We describe practical barriers to deploying these models on edge hardware that go beyond whether the model fits in memory. These include unsupported operators, numerical drift through conversion and limits on single memory allocations.

The rest of the paper is organized as follows. Section 2 covers related work, and Section 3 describes our experiments. Sections 4, 5 and 6 answer the three questions in turn. Section 7 discusses what the results mean, and Section 8 collects the limitations and next experiments.

\section{Background and related work}\label{sec:2}
\subsection{Language models for UAV decisions}\label{sec:2.1}
Ferrag, Lakas and Debbah introduced UAVBench to study how well language models reason about UAV scenarios [1]. They generated 50,000 multiple-choice questions across ten reasoning styles, and released a validated subset of 2,000 of them, which is the set we use (Section 3.1). They then evaluated nearly 40 models on this subset, from small open models such as Llama 3.2 3B to large hosted models such as GPT-5 Chat. The paper reports accuracy separately for each reasoning style, rather than as a single overall score. However, the authors also released their per-question results [2], which match the per-style accuracies in the paper exactly. So we computed each model's overall score from these results.

The best model in their evaluation is Qwen3-235B-A22B, which scores 83.55\% overall. Other large hosted models are close behind, such as ChatGPT-4o (80.35\%), GPT-5 Chat (80.15\%) and Gemini 2.5 Flash (76.75\%). The small models score lower. For instance, Gemma 3n E4B scores 73.25\%, LFM2 2.6B scores 69.75\% and Llama 3.2 3B scores 62.00\%. In their analysis, the authors observe that smaller or lightweight models tend to cluster near 70\%, and conclude that complex UAV reasoning depends on model scale and specialization [1].

Our work builds on theirs in three ways. First, we evaluate a different kind of model, the structured decision models, which they did not test. As we show in Section 4, these models challenge the idea that accuracy on UAVBench simply follows model size. Second, the authors evaluated models through hosted services and focused on the quality of each model's answers. We instead ran every model ourselves, measured its speed and memory use, and scored its answers with a stricter rule (Section 3.1). Finally, we took the most promising models to real edge hardware, and we report both how they performed there and what stopped them from running.

More broadly, language models are being explored for drone autonomy. For instance, Vemprala et al. use ChatGPT to turn high-level instructions into robot code, including aerial navigation tasks [8]. Similarly, TypeFly uses a language model to write short programs that fly a drone from a user's task description [9]. Javaid, Saeed and He survey this growing area [10].

Closer to our setting, SayCan grounds a language model in a robot's available skills [11]. It scores each skill as a candidate next step, combining the language model's judgment of whether the skill helps with a value function that says whether the skill can succeed, rather than generating free text. In other words, it treats the decision as a choice among a fixed set of options, which is the same shape as a UAVBench question. We extend this line of work with controlled local measurements, transfer between inference systems, and the recent open-weight decision models.

\subsection{Structured decision models}\label{sec:2.2}
Structured decision models are an emerging research topic, and this is not the first study of them. Deußer, Sparrenberg and Sifa evaluate the hosted Jev model zero-shot on 37 public datasets, using its choice, yes/no and rubric-scoring question types [12]. Qu, Chen and Labi combine Jev with a language world model for simulated driving and replayed traffic [13]. Delaunay studies the confidence, perturbation sensitivity and explanations of open-weight decision models [14]. What we add is an evaluation on UAVBench, along with local screening, physical measurements on the RB3 and a record of the deployment barriers we hit.

Throughout this paper, a structured decision model means a system that returns a probability distribution over the declared answers. The term describes the interface, not a particular architecture. It does not imply that GLiDE, Liquid d1 and H2O reproduce Jev's undisclosed implementation or training, and we do not evaluate the commercial Jev model itself. The public releases also differ quite a bit from one another. GLiDE supplies a small encoder model (0.4B) and a hybrid model that mixes linear-attention and full-attention layers (0.8B) [4,5]. Liquid d1 supplies its own model, decision computation and calibration [6]. H2O supplies a model along with a released readout and serving implementation [7]. We keep all of these model-specific components. So, although we call every model through the same seven-choice task, their prompts, precision and internal computations are not identical. Similarly, we say ``open-weight'' rather than ``open-source'', because not every release has an unrestricted open-source license.

\subsection{Choosing among options with language models}\label{sec:2.3}
A generative model can also be made to choose among options of answers, and the mechanism to perform this choice influences the end result. Robinson and Wingate show that presenting the question and options together, and asking for the option's letter, works better than scoring each answer on its own, especially for models that link options to their letters well [15]. Zheng et al. show that models tend to prefer particular option IDs, such as ``Option A'', as their answer, which they call selection bias [16]. This matters on UAVBench, where the most common gold letter covers about a quarter of the questions.

Finally, a probability is only useful for decisions if it is calibrated. Intuitively, options given a probability of 0.7 should be right about 70\% of the time. Modern neural networks are often poorly calibrated [17], so a probability returned by a decision model should not be taken at face value.

\subsection{Running language models on edge hardware}\label{sec:2.4}
Running language models on phones, robots and other edge devices is an active area, surveyed by Xu et al. [18]. Most deployments rely on weight quantization to fit the memory budget. For instance, GPTQ compresses weights to 3 or 4 bits with negligible accuracy loss [19], and AWQ targets the same low-bit, weight-only setting for on-device use [20]. In practice, local runtimes such as llama.cpp [21] and Apple's MLX [22] supplied the quantized formats we used.

However, a quantized model is only half the story. The runtime on the target device must also support every operator in the model and produce the same numbers. Yuan et al. show that changing the GPU type, the GPU count or the batch size can change a model's outputs, because small rounding differences compound [23]. We see the same effect when the same model file runs on a Mac and on the RB3.

\section{Experimental setup}\label{sec:3}
Our experiments followed the three questions from Section 1, and ran in two stages. In the first stage, we evaluated all of our models on a Mac, using the full set of 2,000 UAVBench questions. This covered 20 configurations of generative models, as well as the four structured decision models, and it answers our first question. In the second stage, we moved a smaller set of models to the RB3 board. For two of the generative models, we ran the full benchmark on both platforms, so that we could check whether the Mac's results carried over to the board (our second question). We then tried to move all four decision models to the board (our third question). The rest of this section describes the task and how we scored it (Section 3.1), our hardware (Section 3.2), our models (Section 3.3), how we moved them to the RB3 (Section 3.4) and our statistical tests (Section 3.5). Finally, Section 3.6 summarizes every experiment in a single table.

\subsection{Task and scoring}\label{sec:3.1}
We used the validated release of UAVBench, which contains 2,000 multiple-choice questions. We froze it at source revision 4f2baa78a50eea3bb63dffb83565b8f9fb7a7a77 [2], so that every model saw exactly the same questions. The release has 200 questions for each of the benchmark's ten reasoning styles, and every question has seven options (A through G). Note that the authors also released a much larger set of 50,000 questions. We chose the validated 2,000-question subset for two reasons. First, all of the authors' own model results use this subset, so our numbers can be compared directly with theirs (Section 2.1). Second, it keeps the experiments practical. At the board's speed of roughly 4 to 17 seconds per question, the full set of 50,000 questions would take days for each model. Before running any models, we audited the release and found no duplicated questions. We also confirmed that the gold labels and all 14,000 answer options matched the frozen source.

During inference, a model sees only the scenario, the question and the answer options. The gold answers are kept separately and used only afterwards, for scoring. We also did not adapt anything to UAVBench. For instance, we did no training, no prompt tuning based on accuracy, no classifier fitting and no recalibration of probabilities. That said, UAVBench is public, so we cannot rule out that some of its content appeared in a model's original training data. So when we say ``no task-specific training,'' we mean that we did no training of our own.

Our accuracy numbers always use all of the scheduled questions as the denominator, so that nothing is quietly dropped. If a question is longer than a model's token budget, we count it as wrong rather than shortening or removing it. The same goes for failed runs and invalid outputs. To put the numbers in context, a model that picked an option uniformly at random would score 14.29\%. Interestingly, we found that the most common gold letter is C, so a model that always answered C would score 25.30\% without doing any reasoning at all, setting the naïve benchmark.

Generative models and decision models produce different kinds of output, so we score them differently. We ask a generative model to answer with a single letter, and our strict score only accepts an answer in exactly that form. This means the strict score measures whether the model follows the format, as well as whether it picks the right answer. A model might also give the right answer inside a longer sentence (e.g., ``The best option is D because...''). There, we also reported the score from the benchmark's own permissive extractor, which searches the text for a letter. Scoring a decision model is simpler. It returns a probability for each option, and we simply took the option with the highest probability as its answer (if two options tie exactly, we take the earlier letter). We computed this ourselves from the returned probabilities, rather than relying on the ``choice'' field that some serving APIs return for convenience.

\subsection{Hardware}\label{sec:3.2}
We ran our local experiments on an Apple M2 Pro Mac with 32 GiB of unified memory. For the edge platform, we used a Qualcomm RB3 Gen 2 development board. It has a QCS6490 processor and an Adreno 643 GPU, and its runtime exposes about 5.25 GiB of RAM. We accessed the board remotely through the free tier of the Qualcomm Device Cloud. The remote connection was only used to send commands and collect results, and all inference runs were on the board itself. As a result, we excluded network time from our reported timings.

To be clear, we used the Mac as a development machine, not as a simulator of the RB3. We did put memory limits on the Mac to keep the experiments manageable and somewhat reflective. However, these limits do not reproduce the physical board's memory bandwidth, accelerator, power budget or thermal design. So the Mac can tell us which models look promising, but not how they will truly behave once on the board. That gap is exactly what we describe in Section 5.

\subsection{Models}\label{sec:3.3}
We tested two groups of models. The first was a set of generative models, which serves as our baseline. The second was the set of four structured decision models.

Generative models. This set covered 11 checkpoints from six families. These are Qwen2.5 (0.5B, 1.5B, 3B and 7B parameters), SmolLM2 (135M, 360M and 1.7B), LFM2 2.6B, Granite 4 H Micro, Llama 3.2 3B and the text-only version of Gemma 3n E4B. We ran each checkpoint at two precisions, MLX FP16 and MLX 4-bit, whenever it passed our resource checks, which gave us 20 usable configurations. All of the generative runs used the same settings. That is, one request at a time, greedy decoding, a fresh cache for each request, a 2,048-token input limit and a 16-token output limit (which is plenty for a single-letter answer). Finally, our first 4-bit conversion of Gemma produced NaN logits. We excluded it and used a corrected conversion from BF16 weights instead (we discuss this failure in Section 5.1).

Structured decision models. We tested pinned releases of four decision models. These are GLiDE 0.4B and GLiDE 0.8B [4,5], Liquid d1 2B [6] and H2O Lightning 4B [7], which we shorten to Liquid d1 and H2O for clarity. The GLiDE models and Liquid d1 ran in BF16 through Apple MPS. H2O ran as an 8-bit MLX conversion that is linked from H2O's own model card [24]. We served it through H2O's official shim, with the recommended two-way attention and the default profile. In every case, we kept each model's released prompts, label heads and calibration, since these are part of what the publisher actually ships.

Before committing to a full run, we gave each model a short pilot of ten questions (one per reasoning style), which we chose without looking at any model outputs. The pilot only checked that the model ran, and that it gave the same answer when asked the same question twice. It did not require any minimum accuracy. All four models passed. Each then ran on all 2,000 questions with a 4,096-token input limit and no truncation. The references list the exact revision of each model we used. Table 1 summarizes the precision labels we use throughout the paper.

\begin{table}[htbp]\centering\small\setlength{\tabcolsep}{4pt}
\caption{Precision and file-format labels used in this paper.}\label{tab:1}
\begin{tabular}{@{}p{2.5cm}p{7.0cm}p{6.9cm}@{}}\toprule
\textbf{Label} & \textbf{Meaning} & \textbf{Where we use it}\\\midrule
MLX FP16 & 16-bit floating-point weights, run with Apple MLX & Mac generative models\\
MLX 4-bit & MLX affine 4-bit quantization, converted from FP16 weights (BF16 for Gemma) & Mac generative models; MLX resource repeat\\
BF16 & 16-bit bfloat weights, run through Apple MPS & Mac decision study (GLiDE, Liquid d1)\\
MLX q8 & MLX 8-bit conversion of H2O Lightning 4B & Mac decision study (H2O)\\
Q4\_0 GGUF & 4-bit GGUF file, identical on both platforms & Matched Mac--RB3 transfer (Qwen2.5 1.5B, LFM2)\\
Q4\_K\_M GGUF & 4-bit k-quant GGUF file, identical on both platforms & H2O Mac--RB3 diagnostic\\
\bottomrule\end{tabular}\end{table}
\subsection{Moving from the Mac to the RB3}\label{sec:3.4}
Our second and third questions both required running models on the RB3, not just the Mac. So we designed four sets of experiments to compare the two platforms as directly as we could.

Matched transfer. For Qwen2.5 1.5B and LFM2 2.6B, we ran the same official Q4\_0 GGUF file on both platforms, with matched prompts and generation settings. On the Mac, inference used Metal, and on the RB3 it used GenieX with GPU OpenCL. We confirmed that the prompt hashes matched across the two platforms, so both saw exactly the same inputs. Note that the runtime and kernel builds still differed between the two. So this experiment compares two complete deployment systems, rather than isolating the effect of the hardware alone.

Repeated timing. A single timing run can be noisy, so we also repeated our timing measurements. We chose a shared panel of 100 questions and ran six rounds per model on each platform. Each round started a fresh process with its own warmups, and we alternated which model went first. (We note that on the RB3, we also waited before each round until the GPU had cooled to 50 C or below, as a ``cooling check.'') We then took the median time within each round, and we report the median of those six round medians. The timer covered tokenization and generation, but not model loading, warmup or network time. We also logged GPU temperature throughout, since heat can slow a board down.

Numerical diagnostics. We also wanted to understand why some answers changed between platforms. So for each model, we picked 40 questions, 20 where the two platforms had agreed and 20 where they had differed and ran each one three times. Separately, we fed identical saved token IDs to both platforms and compared the logits for the answer letters directly. In this way, any difference we saw could not come from tokenization or prompt handling.

Decision models on the RB3. We tried to move all four decision models to the RB3, but only one of them, H2O, got as far as answering questions on the board. The other three failed at different points along the way, and we describe each failure in Section 6. For every attempt, we first converted the model into a format that the board's runtime could support. We only trusted a conversion if it was faithful to the original model. That is, it had to keep the model's attention, tokenization, pooling, label head and calibration, and it had to reproduce the original model's probabilities on a fixed set of test inputs. If a conversion failed these checks, we reported it as a failed conversion rather than as poor accuracy, since a broken conversion tells us nothing about how well the model itself makes decisions. For H2O, we ran the same GGUF Q4\_K\_M file [25] on both the Mac and the RB3, built from the same runtime source. However, even H2O's first probability check failed, so we tested it on a fixed panel of 100 questions rather than the full benchmark (see Section 5.5).

\subsection{Statistical analysis}\label{sec:3.5}
Most of the differences we report are descriptive. However, for the comparisons that carry our main claims, we ran statistical tests, choosing each one to match how the data were paired.

For LFM2 2.6B against Qwen2.5 7B, both models answered the same 2,000 questions, so we used a paired bootstrap. That is, we resampled the questions 2,000 times, keeping the same number from each model, and recomputed the accuracy difference each time. This gives us an interval for the difference.

For each decision model against the best generative configuration, we took a simpler and more cautious approach. We computed a 95\% interval for the difference that treats the two runs as independent. In practice, two models tend to get the same questions right, and a paired analysis would use this to give a narrower interval. So by ignoring the pairing, our intervals are wider than they need to be, which errs on the side of caution.

For the Mac against the RB3, both platforms answered the same questions, so we used an exact McNemar test [26,27]. Intuitively, a question that both platforms got right (or both got wrong) tells us nothing about which platform is better. So the test looks only at the questions where exactly one platform was correct, and asks whether those cases are split evenly between the two.

We note, importantly, that none of these tests account for the fact that we chose among many candidate models, or for possible overlap between UAVBench and a model's training data.

\subsection{Summary of experiments}\label{sec:3.6}
Table 2 summarizes our five main experiments. Note that they do not all cover the same models or questions. The two Mac studies covered all 2,000 questions. On the RB3, Qwen2.5 1.5B and LFM2 2.6B also had full runs, plus repeated timing runs to measure their latency. H2O had a fixed 100-question diagnostic rather than a full run, since its first probability check failed and the board was slow (Section 3.4). Finally, GLiDE and Liquid d1 contributed evidence about deployment, but no accuracy on the RB3, since we could not get them running on the board (Section 6). We also ran a smaller repeated study of speed and memory on the Mac, which we describe in Section 4.3.

\begin{table}[htbp]\centering\small\setlength{\tabcolsep}{4pt}
\caption{The five main studies, their coverage and their main controls.}\label{tab:2}
\begin{tabular}{@{}p{3.2cm}p{6.1cm}p{7.1cm}@{}}\toprule
\textbf{Study} & \textbf{Models and coverage} & \textbf{Main controls}\\\midrule
Mac generative models & 11 checkpoints; 20 usable conditions; 2,000 questions each & MLX FP16 or MLX 4-bit; 2,048-token input limit; 8 GiB allocator allowance\\
Mac decision models & GLiDE 0.4B and 0.8B, Liquid d1 2B, H2O Lightning 4B; 2,000 each & Native decision code; BF16 or MLX q8; 4,096-token input limit\\
Matched full transfer & Qwen 1.5B and LFM2 2.6B; 2,000 per model and platform & Identical Q4\_0 GGUF files and prompts; Metal versus OpenCL\\
Repeated transfer timing & Two models; six rounds of 100 questions on each platform & Fresh processes; separate warmup; alternating model order\\
H2O transfer diagnostic & Same fixed 100 questions on Mac and RB3 & Identical Q4\_K\_M GGUF file and runtime source; 1,024-token context\\
\bottomrule\end{tabular}\end{table}
\section{Small decision models as candidates}\label{sec:4}
\subsection{Decision models score highest without task-specific training}\label{sec:4.1}
Table 3 gives the full Mac results, and Figure 2 plots accuracy against speed. Most interestingly, the decision models are strong candidates. Even the smallest, GLiDE 0.4B, scores 66.30\%, which is close to the 66.15\% of Qwen2.5 7B in MLX 4-bit, a much larger generative model. The other three decision models all beat the best generative configuration, Gemma 3n E4B in MLX 4-bit, which scores 70.20\%.

\begin{figure}[H]\centering\includegraphics[width=\linewidth]{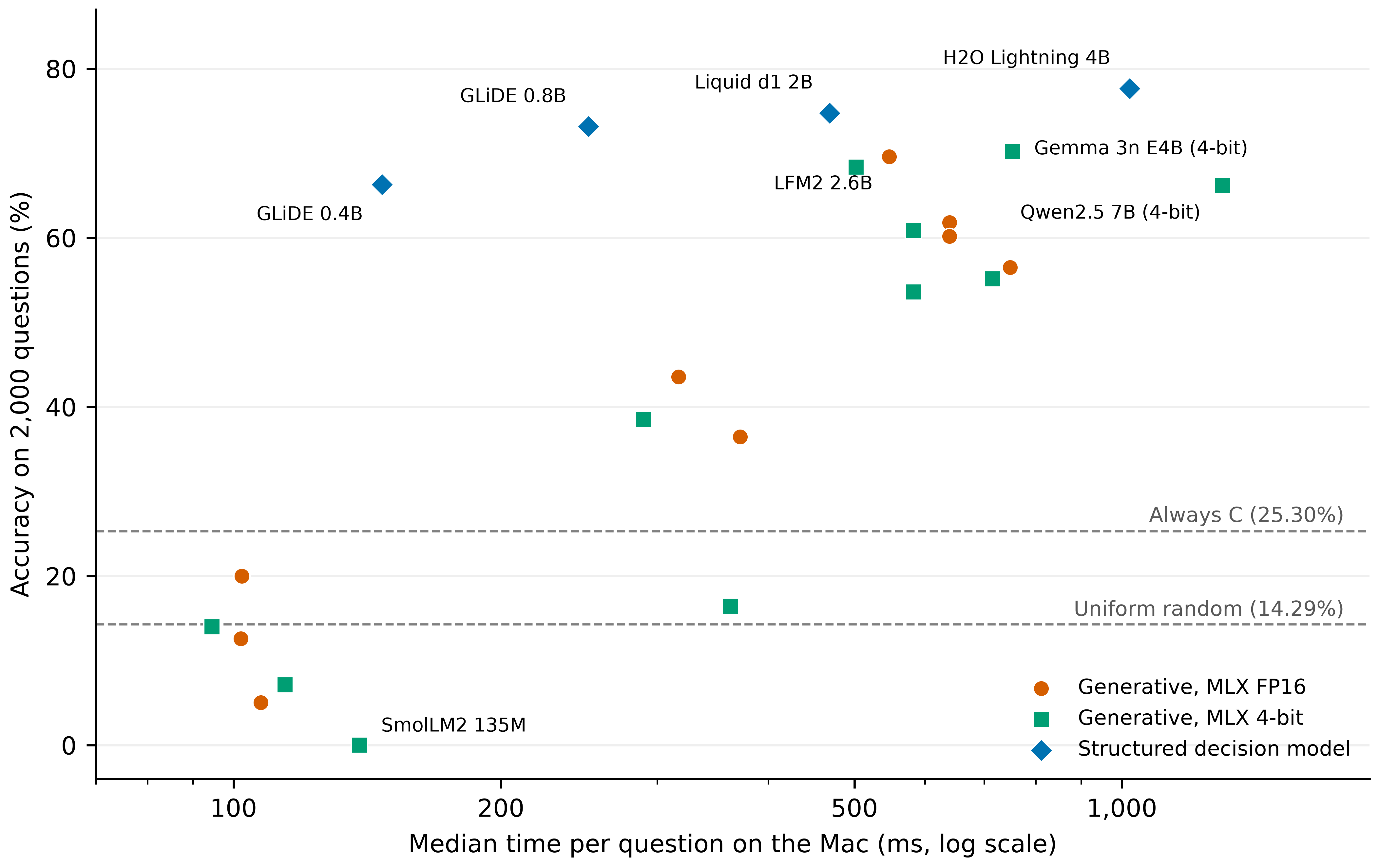}\caption{Accuracy against median time per question on the Mac for all 24 configurations in Table 3. The decision-model times include prompt construction and tokenization, so if anything they are overstated relative to the generative times. The dashed lines show uniform random choice and always answering C.}\label{fig:2}\end{figure}
To put these gaps in concrete terms, H2O was 7.45 points ahead of Gemma, which means it answered 149 more of the 2,000 questions correctly. Liquid d1 was 4.55 points ahead (91 more questions), and GLiDE 0.8B was 2.95 points ahead (59 more questions). Of course, part of any gap could simply be luck involving which questions happen to be in the benchmark. Therefore, for each gap we also computed a 95\% interval, which gives the range of differences that are consistent with our data (Section 3.5). Intuitively, if that range includes zero, we cannot rule out that the two models are equally good. For H2O, the range was 4.74 to 10.16 points, and for Liquid d1 it was 1.79 to 7.31 points, so both are clearly ahead of Gemma. GLiDE 0.8B's range was 0.16 to 5.74 points, so it is also ahead, but only slightly, since the low end is barely above zero. Finally, recall that we built these intervals to be cautious. A paired analysis would likely give narrower ranges, which would only make the decision models' lead clearer.

\clearpage\begingroup\fontsize{9}{11}\selectfont\setlength{\tabcolsep}{3pt}
\begin{longtable}{@{}>{\raggedright\arraybackslash}p{3.0cm}>{\raggedright\arraybackslash}p{1.8cm}rrrrr@{}}
\caption{Full UAVBench results for all 24 configurations on the Mac. Each configuration is scored over all 2,000 questions.}\label{tab:3}\\
\toprule
Model & Precision & Correct / 2,000 & Accuracy (\%) & Permissive (\%) & Median time (ms) & Peak alloc. (GiB)\\\midrule
\endfirsthead
\multicolumn{7}{l}{\textit{Table \thetable\ continued}}\\\toprule
Model & Precision & Correct / 2,000 & Accuracy (\%) & Permissive (\%) & Median time (ms) & Peak alloc. (GiB)\\\midrule
\endhead
\bottomrule\endfoot
SmolLM2 135M & MLX FP16 & 101 & 5.05 & 13.50 & 107.3 & 0.625\\
SmolLM2 135M & MLX 4-bit & 0 & 0.00 & 12.50 & 138.6 & 0.343\\
SmolLM2 360M & MLX FP16 & 252 & 12.60 & 14.55 & 101.9 & 1.285\\
SmolLM2 360M & MLX 4-bit & 143 & 7.15 & 12.50 & 114.3 & 0.638\\
SmolLM2 1.7B & MLX FP16 & 729 & 36.45 & 36.85 & 371.7 & 3.653\\
SmolLM2 1.7B & MLX 4-bit & 329 & 16.45 & 16.55 & 362.8 & 1.624\\
Qwen2.5 0.5B & MLX FP16 & 400 & 20.00 & 20.00 & 102.2 & 1.439\\
Qwen2.5 0.5B & MLX 4-bit & 280 & 14.00 & 14.00 & 94.5 & 0.673\\
Qwen2.5 1.5B & MLX FP16 & 871 & 43.55 & 43.55 & 316.8 & 3.235\\
Qwen2.5 1.5B & MLX 4-bit & 770 & 38.50 & 38.50 & 289.7 & 1.546\\
Qwen2.5 3B & MLX FP16 & 1,236 & 61.80 & 61.80 & 639.5 & 6.055\\
Qwen2.5 3B & MLX 4-bit & 1,072 & 53.60 & 53.60 & 583.3 & 2.328\\
Qwen2.5 7B & MLX 4-bit & 1,323 & 66.15 & 66.15 & 1,298.3 & 4.568\\
LFM2 2.6B & MLX FP16 & 1,392 & 69.60 & 69.60 & 547.1 & 5.090\\
LFM2 2.6B & MLX 4-bit & 1,367 & 68.35 & 68.35 & 502.1 & 2.037\\
Granite 4 H Micro & MLX FP16 & 1,130 & 56.50 & 56.50 & 748.3 & 7.888\\
Granite 4 H Micro & MLX 4-bit & 1,103 & 55.15 & 55.15 & 714.9 & 3.915\\
Llama 3.2 3B & MLX FP16 & 1,204 & 60.20 & 60.20 & 639.6 & 6.313\\
Llama 3.2 3B & MLX 4-bit & 1,218 & 60.90 & 60.90 & 582.6 & 2.330\\
Gemma 3n E4B text & MLX 4-bit & 1,404 & 70.20 & 70.20 & 752.9 & 4.733\\
\midrule
GLiDE 0.4B & BF16 & 1,326 & 66.30 & -- & 147.0 & --\\
GLiDE 0.8B & BF16 & 1,463 & 73.15 & -- & 250.9 & --\\
Liquid d1 2B & BF16 & 1,495 & 74.75 & -- & 468.7 & --\\
H2O Lightning 4B & MLX q8 & 1,553 & 77.65 & -- & 1,020.6 & --\\
\end{longtable}\endgroup
Scoring. The first 20 rows are generative models, scored with the strict single-letter rule. The permissive column gives the benchmark extractor's score. The last four rows are decision models, scored by the most probable option. Overflows and failures count as incorrect, and the numerically invalid Gemma conversion is excluded.

Precision. Labels follow Table 1. The Gemma 4-bit row is converted from BF16 weights, and the other 4-bit rows from FP16. Peak allocation is the MLX peak, not total RAM, and is reported for the generative models only.

Timing. Times are single-pass medians over executed requests, with loading and warmup excluded. For generative models they cover prefill and generation. For decision models they also cover prompt construction, tokenization and the probability readout, plus a local HTTP call for H2O. Because the timing scopes differ, the two groups are not strictly comparable on speed.

These results also put the decision models in a broader context. Recall from Section 2.1 that the UAVBench authors found smaller models tend to cluster near 70\%, while the best large model they tested, Qwen3-235B-A22B, scores 83.55\%. H2O, a 4B model running on a Mac, scores 77.65\%. That is within about six points of the best model in their evaluation, and slightly above large hosted models such as Gemini 2.5 Flash (76.75\%). Their numbers use the benchmark's permissive extractor and hosted versions of each model, so this comparison is indicative rather than exact. Even so, it suggests that on this benchmark, a small model with the right interface can come close to much larger models.

While the structured models produced interesting results, we do want to caution what the comparison shows. First, we compared complete packages, rather that only isolating the models, and many aspects can differ between the two groups besides the decision-model design itself. For instance, GLiDE 0.8B ran at full BF16 precision, while Gemma ran in 4-bit, and the two also used different prompts and different inference software. Any of these differences could account for part of the gap. Our result can help a practitioner decide among various released models for their task, however it does not explicitly show that the decision-model design is better in itself. We describe an experiment that would separate these effects in Section 7.1.

\subsection{Decision models offer different accuracy and speed choices}\label{sec:4.2}
The decision models also offer a range of choices between accuracy and speed. H2O was the most accurate at 77.65\%, with a median call of 1.021s, while GLiDE 0.8B reached 73.15\% at 0.251s, about four times faster. Liquid d1 sits between them, at 74.75\% and 0.469s. Finally, GLiDE 0.4B was the fastest of the set, at 0.147s, but it scored lowest at 66.30\%. We note, however, these are single passes with different precisions and implementations, so we caution to use them as a ranking rather than relying on the explicit numbers themselves.

\subsection{What the generative models show}\label{sec:4.3}
Although the generative models were mainly our baseline, they still teach us two things about choosing a small model. The first is that, within a family, bigger models do better. For instance, among the Qwen2.5 models in MLX 4-bit class, strict accuracy rises from 14.00\% at 0.5B to 38.50\% at 1.5B, 53.60\% at 3B and 66.15\% at 7B. The SmolLM2 models also improve with size. So, even though the smallest models need very little memory, they are weak candidates for this task. Note, however, that a family's releases can differ in how they were trained as well as in their size. So this trend tells us which releases to pick, but it does not isolate the effect of parameter count on its own.

The second lesson is that a smaller model from a different family can match a larger one at a much lower cost. For instance, LFM2 2.6B in MLX 4-bit scores 68.35\%, slightly higher than the 66.15\% of Qwen2.5 7B, a model with nearly three times as many parameters. The 2.20-point gap itself is not meaningful, since its paired bootstrap interval runs from -0.05 to +4.40 points, which includes zero. In other words, the safest reading is that the two models are about equally accurate. The difference in cost, however, is large. To measure it reliably, we ran a separate study on the Mac, with six rounds of 100 questions per model (Table 4). LFM2 was about 2.57 times faster than Qwen2.5 7B and used about 55\% less memory at its peak. Gemma 3n E4B, the best generative model, was slightly more accurate than LFM2, but it took about 1.5 times as long per question and needed more than twice the memory. Finally, as a check on the repeated runs themselves, all 1,800 timed requests gave the same answers as the original full runs.

\begin{table}[htbp]\centering\small\setlength{\tabcolsep}{4pt}
\caption{Repeated MLX resource study, with six rounds of 100 questions per model. Times are medians of the six round medians. Allocation is the peak MLX allocation, not total RAM.}\label{tab:4}
\begin{tabular}{@{}lrrrr@{}}\toprule
\textbf{Model in 4-bit} & \textbf{Full accuracy} & \textbf{Median time} & \textbf{Round range} & \textbf{Peak allocation}\\\midrule
LFM2 2.6B & 68.35\% & 505.5 ms & 503.9--507.4 ms & 2.035 GiB\\
Gemma 3n E4B text & 70.20\% & 756.1 ms & 753.2--761.4 ms & 4.726 GiB\\
Qwen2.5 7B & 66.15\% & 1,301.7 ms & 1,298.7--1,304.4 ms & 4.562--4.563 GiB\\
\bottomrule\end{tabular}\end{table}
As a sanity check, we also compared our results with the original UAVBench results on the same 2,000 questions. The authors' released per-question results give overall scores of 66.05\% for Qwen2.5 7B, 69.75\% for LFM2 2.6B, 57.80\% for Granite 4 H Micro, 62.00\% for Llama 3.2 3B and 73.25\% for Gemma 3n E4B [2]. (The paper itself reports these models only by groups of reasoning styles [1].) All five are within about three points of our own scores in Table 3, which suggests that our setup behaves sensibly. However, our numbers are not an exact reproduction. The model revisions and precisions the authors served differ from ours, and their raw generations are not available, so we could not rescore them with our strict rule.

\section{Screening on the Mac before testing on the board}\label{sec:5}
Our second question is whether a local machine, such as a Mac, can be used to pick promising candidates before testing them on the board. In short, the answer is yes, but with some care. First, before trusting any local score, we have to check that the score is valid (Section 5.1). Once it is, the Mac ranks candidates well (Section 5.2). However, it does not predict which individual questions a model will get right (Section 5.3), how fast the model will run on the board (Section 5.4) or, for decision models, the exact probabilities the model returns (Section 5.5).

\subsection{Check validity before trusting a local score}\label{sec:5.1}
Before using a local score to rank models, we need to make sure that it measures what we think it measures. In our experiments, we found three ways that a score could be misleading.

The first is quantization. Quantization stores a model's weights with fewer bits (here, 4 bits instead of 16), so that the model uses less memory. Ideally, this costs little accuracy, and for some models that is true. For instance, moving from MLX FP16 to MLX 4-bit costs LFM2 only 1.25 points (69.60\% to 68.35\%), and Granite only 1.35 points. Other models suffer much more. Qwen2.5 3B loses 8.20 points, and SmolLM2 1.7B loses 20.00. Llama 3.2 3B actually gains a little, from 60.20\% to 60.90\%, but a gap this small is not evidence that quantization helps. The lesson is that quantization has to be checked separately for each model, because a setting that saves memory can also remove much of a model's usefulness.

The second is how we read the model's answer. As described in Section 3.1, our strict rule only accepts a single letter, while the benchmark's permissive extractor searches the model's text for a letter. Most of the time, the two agree. Indeed, for every family except SmolLM2, the two scores match exactly (Table 3). The very small models are different. For instance, SmolLM2 135M in MLX FP16 scores 13.50\% with the permissive extractor, but only 5.05\% with the strict rule, and its 4-bit version scores 12.50\% and 0.00\%, respectively. When a model writes a sentence instead of a single letter, the permissive extractor can still find a letter in it, and sometimes that letter happens to be right. So the strict score measures whether a model follows the requested format, as well as whether it picks the right answer. In practice, this choice mainly changes how we read the weakest models. Note that we do not have the original UAVBench raw outputs, so we could not rescore those models with our strict rule. As a result, we make no claim about how their published scores would change.

The third is numerical failure. Our first 4-bit conversion of Gemma, made from FP16 weights, produced NaN (``not a number'') logits on all six of the diagnostic requests we sampled. In other words, a calculation inside the model had broken down, so its output no longer reflected any real reasoning. However, a broken run like this can still return something that looks like an answer. So without checking, it would simply look like a weak model. When we instead converted Gemma from BF16 weights, the logits stayed finite on the diagnostics and on all 1,999 executed requests, and the model scored 70.20\%, the best generative result. For the other configurations, we only checked for finite logits on a sample of requests. So those checks tell us about the sampled requests, rather than guaranteeing that every request was valid.

\subsection{Mac accuracy ranks candidates for board testing}\label{sec:5.2}
Since testing on the board is slow and costly, the most useful thing a Mac can do is tell us which models are worth testing there. To check whether it can, we ran Qwen2.5 1.5B and LFM2 2.6B on the full benchmark on both platforms, using the same Q4\_0 file on each (Table 5). Overall accuracy was very close. Qwen2.5 1.5B scored 41.10\% on the Mac and 41.05\% on the RB3, a difference of a single question. LFM2 scored 69.30\% on the Mac and 68.30\% on the RB3, which means the RB3 answered 20 fewer questions correctly.

We then tested whether these small drops are real, using the McNemar test described in Section 3.5. Neither is significant (p = 1.0 for Qwen2.5 1.5B and p = 0.11 for LFM2). In other words, gaps of this size could easily arise by chance, so we have no evidence that moving to the board made either model less accurate.

More importantly for screening, the ranking held. LFM2 was far ahead of Qwen2.5 1.5B on the Mac, and it stayed far ahead on the RB3. This supports using the Mac to discard weak candidates and identify promising ones, before spending time on the board. Of course, two models is a small sample. Also, a one-point drop like LFM2's could reorder two candidates whose scores are close, so the Mac is better at separating strong models from weak ones than at choosing between near-equals.

\begin{table}[htbp]\centering\small\setlength{\tabcolsep}{4pt}
\caption{Matched Q4\_0 transfer on the full benchmark. Accuracy uses all 2,000 questions. Answer agreement uses the 1,999 executed questions, since one overlong question counts as incorrect on both platforms.}\label{tab:5}
\begin{tabular}{@{}lrrrr@{}}\toprule
\textbf{Model} & \textbf{Mac accuracy} & \textbf{RB3 accuracy} & \textbf{Identical answers} & \textbf{Full RB3 median}\\\midrule
Qwen2.5 1.5B & 41.10\% & 41.05\% & 1,943 / 1,999 & 3.942 s\\
LFM2 2.6B & 69.30\% & 68.30\% & 1,775 / 1,999 & 6.306 s\\
\bottomrule\end{tabular}\end{table}
\subsection{Similar accuracy can hide different answers}\label{sec:5.3}
Close overall scores do not mean that the two platforms gave the same answers. In fact, Qwen2.5 1.5B changed its answer between the Mac and the RB3 on 56 of the 1,999 executed questions, and LFM2 changed on 224, which is about one question in nine. For Qwen2.5 1.5B, the Mac was right and the RB3 wrong on 13 of these questions, and the reverse happened on 12. For LFM2, the counts were 79 and 59. On the remaining changed questions, both platforms gave a wrong answer, just a different one. Most interestingly, the gains and losses largely cancel each other out, so the overall scores stay close, even though many individual answers changed.

A natural question is whether these changes are just random, that is, whether running the same question again on the same platform would give a different answer. To find out, we reran 20 of the changed questions per model three times on each platform (Section 3.4). The changes were not random. All 20 disagreements reappeared in every repetition, and within each platform, the scores for the seven answer letters were identical across repetitions on all 40 questions we tested per model. We then fed the exact same token IDs to both platforms, which rules out any difference in how the prompt was tokenized or handled. The two platforms still picked different answers. As a final check, the answer the board gave from these token IDs matched its normally generated answer on all 40 questions.

In other words, the Mac and the RB3 each compute a consistent result from the same model file, but the two results are not the same. This matches what Yuan et al. found when they changed GPU configurations [23], where small differences in how numbers are rounded add up to a different output. To be clear, we did not trace our differences to a specific kernel or rounding step.

The practical lesson is that using the same model file, prompt and seed does not guarantee the same decision on a different computer. A Mac can tell us that a model is broadly promising. However, testing on the actual hardware still has to look at which answers changed, not just at the overall score. This matters most for high-consequence decisions. For instance, a drone running LFM2 on the board could make a different call than the Mac version in a situation like the one in Figure 1.

\subsection{Mac latency does not predict board latency}\label{sec:5.4}
As expected, the board was much slower than the Mac (Table 6). Qwen2.5 1.5B took a median of about 0.3 s per question on the Mac, but about 4.0 s on the RB3, roughly 12.7 times longer. LFM2 took about 0.5 s on the Mac and about 6.3 s on the RB3, again roughly 12.1 times longer. These measurements were also very stable. Across the six rounds, the median times on the board stayed within a few milliseconds of each other, and both models gave the same answer to each of the 100 questions in every round.

Since both models slowed down by a similar amount, it is tempting to treat ``about 12 times slower'' as a rule for converting Mac timings into board timings, but we caution against this. The slowdown depends on the device, the runtime and the model. For instance, H2O slowed down by about 17.3 times on the same board (Section 5.5). So Mac timing can tell us which models are faster, but only relatively. The actual speed on the board has to be measured there.

\begin{table}[htbp]\centering\small\setlength{\tabcolsep}{4pt}
\caption{Repeated matched timing in milliseconds. Each cell summarizes six rounds of 100 questions. Ranges span the six round medians.}\label{tab:6}
\begin{tabular}{@{}lrrrr@{}}\toprule
\textbf{Model} & \textbf{Mac median} & \textbf{Mac range} & \textbf{RB3 median} & \textbf{RB3 range}\\\midrule
Qwen2.5 1.5B & 313.7 & 312.5--314.0 & 3,979.2 & 3,977.2--3,980.9\\
LFM2 2.6B & 525.8 & 525.7--526.4 & 6,338.4 & 6,333.4--6,339.2\\
\bottomrule\end{tabular}\end{table}
We also tracked GPU temperature, since a board that gets too hot can slow itself down to cool off (known as throttling). The board has two GPU sensors. During the timed rounds, they read 55.4--69.8 C and 58.1--79.1 C for Qwen2.5 1.5B, and 55.8--74.0 C and 56.6--83.7 C for LFM2. The full 2,000-question runs got hotter, reaching peaks of 75.2 C and 84.9 C for Qwen2.5 1.5B, and 80.6 C and 91.1 C for LFM2. However, the full runs and the timed rounds differed in length and in starting temperature, so we cannot use these readings to say which model runs hotter. Also, a high temperature on its own does not prove that the board throttled. We saw no sign of throttling in the timed rounds, since their median times stayed nearly constant from round to round.

\subsection{H2O transfers accuracy but not probabilities or speed}\label{sec:5.5}
H2O was the only decision model that we got running on the RB3 (Section 3.4). Since its first probability check had failed, and since the board is slow, we tested it on a fixed panel of 100 questions rather than the full benchmark (see Table 7 below). On these questions, the answers transferred well. Both platforms got 80 out of 100 right, and they gave the same answer on 97 questions. On the remaining three, each platform got one question right that the other got wrong, and on the third they gave different wrong answers.

The probabilities did not transfer as well. This matters for a decision model, because its probabilities are a large part of what it offers. For instance, a system might only act on a choice when the model gives it a probability above some threshold. We found that the probabilities for at least one option differed by more than 0.01 between the platforms on 32 of the 100 questions. The largest difference was 0.1718, which is easily enough to push an option across a threshold such as 0.5. H2O's publisher supplies its own probe questions to check a conversion, and these passed. However, the publisher's probes allow differences up to 0.05, which is looser than our 0.01 tolerance. So the probability mismatch we found in our first check remains unresolved.

\begin{table}[htbp]\centering\small\setlength{\tabcolsep}{4pt}
\caption{H2O on the same fixed 100 questions on both platforms, using the same Q4\_K\_M GGUF file. These are diagnostic results, not full-benchmark accuracy.}\label{tab:7}
\begin{tabular}{@{}lrr@{}}\toprule
\textbf{Measure} & \textbf{Mac} & \textbf{RB3}\\\midrule
Correct answers & 80 / 100 & 80 / 100\\
Median response time & 0.967 s & 16.762 s\\
Mean response time & 0.980 s & 16.827 s\\
95th percentile & 1.150 s & 19.809 s\\
Timed GPU sensor 0 range & Not measured & 47.3--62.4 C\\
Timed GPU sensor 1 range & Not measured & 46.9--70.9 C\\
\bottomrule\end{tabular}\end{table}
Speed transferred worst of all. H2O took a median of 0.967 s per question on the Mac, but 16.762 s on the RB3, about 17.3 times longer. The slowest questions took longer still, since one question in twenty took nearly 20 s on the board (the 95th percentile was 19.809 s). As a rough planning estimate, a full 2,000-question run at the board's mean time would take about 9.35 hours. Note that we did not measure this. Our panel's inputs were also at most 571 tokens long, so longer questions remain untested. Similarly, we do not claim that the panel's 80\% accuracy would hold across the full benchmark.

\section{Software and resource limits determine deployment}\label{sec:6}
Our third question is what gets in the way of actual deployment of these models. In short, the Mac results tell us which models are worth deploying, but not which ones can be deployed. Of our four decision models, three never answered a single question on the board, and the fourth, H2O, needed several workarounds to do so. In this section, we describe what went wrong for each of them, since each failure points to a different hurdle.

\subsection{GLiDE 0.4B: a graph the runtime cannot finalize}\label{sec:6.1}
GLiDE 0.4B was the most promising candidate for the board, since it was the fastest decision model on the Mac and still reasonably accurate. Its export also passed all of our local checks. However, it failed on the board itself. Before running a model, the board's runtime (here, its QNN GPU path) has to compile the exported model into a form the GPU can execute, a step called graph finalization. GLiDE 0.4B failed at this step. We tried two adaptations that change how the model represents its indices and masks (one using integer indices and one using boolean masks). Both kept the model's probabilities unchanged on our pilot questions, but neither fixed the finalization failure. To make sure the GPU path worked at all, we also ran a small matrix multiplication on the board with CPU fallback turned off, and it succeeded. This shows that the GPU backend can handle small operations. However, it does not show that the backend supports the full model. In the end, we could not complete any benchmark run of GLiDE 0.4B on the RB3.

\subsection{GLiDE 0.8B: an unsupported operator}\label{sec:6.2}
GLiDE 0.8B failed even earlier than GLiDE 0.4B, since it never made it onto the board at all. To run a model on the board, we first had to export it, which means converting each of the model's mathematical operations (called operators) into a format that the board's runtime understands. If the export tool does not support an operator, the export stops. GLiDE 0.8B's recurrent attention layers use an operator called a triangular solve, and the ONNX export path we chose does not support it. We could have replaced GLiDE 0.8B's hybrid attention with an ordinary causal decoder, which the exporter does support. However, that would have changed the model, so we would no longer have been testing GLiDE 0.8B. We therefore stopped. To be clear, this is a limit of the export route we tried, not proof that GLiDE 0.8B cannot run on any edge accelerator.

\subsection{Liquid d1: a failed tolerance check}\label{sec:6.3}
Liquid d1 also hit an export problem, but one we could work around. Its attention layers use a grouped-query attention call (enable\_gqa=True) that the exporter did not support. So we rewrote that call in a form the exporter could handle. To check that the rewrite did not change the model, we compared it against the original version on our ten pilot questions. Both the rewritten model and the exported version made the same choices as the original on all ten. However, the probabilities were not quite the same. On two of the questions, the exported probabilities differed from the original by more than our predefined tolerance of 0.001, with a largest difference of 0.00207. We therefore counted this as a failed check and did not run Liquid d1 on the board. These differences are small, and the choices did match. However, matching choices on only ten questions is not a good enough reason to quietly loosen a rule that we set in advance based on probabilities

\subsection{H2O: a limit on single allocations}\label{sec:6.4}
H2O reached the board, but only after three fixes. First, we made sure the exported model kept H2O's attention behavior. Second, the model file declared an extra block that H2O does not use, and this declaration stopped the runtime from loading it. We removed the declaration, changing only the file's metadata, and we verified that every byte of the model's weights was unchanged. Third, and most important, the RB3 runtime would not hand out more than 256 MiB in any single GPU memory allocation. H2O has one tensor that is about 497 MiB. That tensor holds the word embeddings, which the model shares between its input and output layers. Since it could not fit in a single GPU allocation, it had to stay on the CPU. As a result, H2O ran with part of its work on the CPU and the rest on the GPU (through OpenCL), so it is not a GPU-only result.

Note that this 256 MiB limit is not the same as the board's total memory (about 5.25 GiB), and it is also not the same as the memory figure the GPU driver reports. In fact, H2O's buffers on the GPU totaled about 2,004 MiB, far more than 256 MiB, because they were split across many smaller allocations. As an analogy, a parking lot can have plenty of total space but no single spot big enough for a bus. So the question ``will it fit?'' has several different answers, depending on which limit we ask about.

\subsection{RB5: a driver and runtime problem}\label{sec:6.5}
We also tried an RB5 board, to see whether the same software would run there. The runtime found the board's Adreno GPU, but it then hit an interface problem when querying the GPU's subgroup support, followed by a failure to compile its GPU kernels. As a result, not a single benchmark question completed. This tells us about the combination of driver and runtime installed on that board. It does not tell us anything about the RB5's accuracy, or about whether its hardware could run these models with different software.

\subsection{What makes a deployment candidate}\label{sec:6.6}
Together, these cases suggest a broader definition of a deployment candidate. Of course, a candidate must perform well enough on the task. However, it must also clear several other hurdles, and each of our failed models tripped on a different one. It must fit within the board's memory limits, including the limit on any single allocation, which was the problem for H2O. It must keep its computation intact through conversion, which was the problem for Liquid d1. Finally, it must use only operators that the target runtime supports, which was the problem for GLiDE 0.8B. A model's parameter count and file size describe only part of this picture. We report these failed attempts because, in practice, they decided which of our promising local models could be tested on the board at all.

\section{Discussion}\label{sec:7}
\subsection{What the decision-model results add}\label{sec:7.1}
Our clearest new result is that small open-weight decision models are strong candidates for text-based UAV decisions. For instance, GLiDE 0.8B reached 73.15\% accuracy with a call of about a quarter of a second on the Mac, and H2O reached 77.65\%. However, our deployment results add an important twist, which is that the best candidate on paper is not necessarily the easiest to deploy. GLiDE 0.8B looked very promising on the Mac, but we could not get it onto the RB3 through our export route. H2O did run on the RB3, but only with part of the model on the CPU, with much longer response times and with probabilities that drifted from the Mac's. So whether a model can actually run on the target, and run faithfully, has to be part of choosing it, not something checked afterwards.

We also have not yet isolated why the decision models score so well. As we noted in Section 4.1, the two groups differ in training, prompts, precision, attention and inference software. One difference stands out, though. Our generative baselines write free text, while the decision models are forced to pick one of the seven options. A fair follow-up would give the generative models the same advantage. That is, rather than letting a generative model write an answer, we would read off the probability it assigns to each of the letters A through G, and take the most likely letter as its answer, in the spirit of Robinson and Wingate [15]. For instance, on the question in Figure 1, Gemma would no longer write a sentence about wind and altitude. It would simply give us seven probabilities, just as a decision model does. If the decision models still lead after this change, then the advantage comes from the models themselves. If the gap closes, then much of it came from the interface.

\subsection{A practical local screening procedure}\label{sec:7.2}
Our results suggest a staged procedure for screening models locally before deploying them. First, freeze the questions, labels, token limits and scoring rules before comparing any models, so that every model is measured the same way. Second, check that each run is numerically valid and fits the resource budget before reading its accuracy, since, as Section 5.1 showed, a broken run can look like a weak model. Third, use the local results to shortlist a few candidates whose accuracy and speed suit the mission. Finally, build a faithful version of each candidate on the target device, repeat the numerical checks on a fixed set of inputs, and only then collect accuracy and timing results.

When validating on the target hardware, we also recommend reporting how often the answers agree across platforms, not just the overall accuracy. As our Qwen2.5 1.5B and LFM2 results showed, the overall scores can match closely even when many individual answers change. For models that return probabilities, compare the probabilities as well as the choices. For timing, repeat the measurements on a shared set of questions, state exactly what the timer includes, and log the temperature. Finally, if the version on the target differs from the local one in precision or in where it runs (for instance, part of the model on the CPU), treat it as a new condition and build a matching local reference for it, as we did for H2O.

\subsection{What an onboard claim would need}\label{sec:7.3}
For a drone, response time, temperature and power all matter, but they answer different questions, and we measured only the first two. Power matters because a drone's battery limits how long it can fly. We kept the RB3's raw power readings, but we have not checked their units or exactly what they measure. So we report no energy figures, and we make no claim about how these models would affect flight time.

Timing is the other gap. The use case we have in mind is an occasional high-level mission decision, not flight control. Even so, the fastest generative model on the RB3 took about four seconds per decision, and H2O took about seventeen. Whether that is fast enough depends on how often the mission needs a decision, and on what the drone does while it waits. For instance, in the Figure 1 scenario, a drone waiting seventeen seconds for an answer would keep flying its current plan in strong wind, on a limited battery. So a real deployment study should measure how often the model picks the right action in time, under stated deadlines and sustained use, and while the board is also busy with other work such as perception or navigation.

\section{Limitations and next experiments}\label{sec:8}
Rather than repeating caveats throughout the paper, we collect the main ones here.

 The benchmark. UAVBench describes each situation in text, with the drone's state given directly (e.g., its altitude and battery level). In a real drone, that state would come from sensors, which can be noisy or fail. The benchmark also does not test what happens after the model acts, or how the drone flies when its decisions feed back into the next situation (closed-loop flight). Similarly, the benchmark's labels tell us which answer its authors marked as correct. They do not prove that the answer would be safe in a real flight. For instance, the gold answer to the question in Figure 1 is D, but the label alone does not show that dropping to 60 m would be safe in that exact airspace. Finally, accuracy varies a lot across the ten reasoning styles. For example, the corrected Gemma model scores 47.0\%, 51.0\% and 95.0\% on styles 1, 3 and 6, respectively. So a good overall score can hide serious weaknesses in particular kinds of reasoning.

The comparisons. As discussed in Sections 4.1 and 7.1, the generative and decision models differ in prompts, precision, runtimes and what our timers include. So none of our comparisons isolates the effect of model design or model size. Our set of candidate models is also broad, but not exhaustive. In addition, model revisions and precisions varied across stages of the study, and we added some of our numerical checks only after the Gemma failure. When we rejected a model for exceeding our memory budget, that means it failed under the allowance we chose, not that it could never run on a Mac. Finally, we have not yet measured calibration. Until we do, the probabilities that the decision models return should not be read as true measures of confidence. That is, a model that gives an option a probability of 0.9 is not necessarily right 90\% of the time.

The measurements. Our repeated timing comes from single sessions on one Mac and one RB3, and we did not fully control the room temperature or other background work on those machines. We did see that the Mac was already using some swap memory (disk space used as overflow when memory runs short). However, we did not record whether the model itself was being swapped in and out, so we cannot rule out that this affected our timings. On the RB3, our cooling check ran before the warmup requests, so the timed rounds did not all start at the same temperature. Our H2O results come from an exploratory panel of only 100 questions. Finally, GLiDE and Liquid d1 produced no results on the board through the routes we tried. This should not be read as a statement about how efficient those models are.

Next experiments. Our first priority is to get GLiDE 0.8B or Liquid d1 running faithfully on an edge device. Trying a second kind of hardware, such as an NVIDIA Jetson, would also tell us whether the failures we saw are specific to the Qualcomm software we used. Beyond deployment, we plan to score the generative models by the probabilities they give to each option letter, as described in Section 7.1. We also plan to measure calibration and abstention on questions held out from tuning, where abstention means the model declines to answer when it is unsure. Finally, before making any claims about energy use or real deployment, we will need validated power measurements taken at the board's power input, longer thermal trials, and tests in which the board also runs the kind of work a real drone would need, such as perception and navigation.

\section{Conclusion}\label{sec:9}
In this paper, we asked three questions about running small language models on a drone's computer. First, we asked whether small structured decision models are good candidates for UAV mission decisions. We found that they are. All four open-weight releases we tested did well on UAVBench without any training of our own, and three of them beat every generative configuration we tried, with the best, H2O, reaching 77.65\%. Second, we asked whether a Mac can screen candidates before testing on the board. It can, as a first filter, since it ranked our two matched models the same way the RB3 did. However, the same model file still changed some of its answers when it moved to the board, and it ran roughly 12 to 17 times slower there. Third, we asked what gets in the way of deployment. When we tried to port the decision models, three of the four never answered a question on the board, for three different reasons, none of which model size would have predicted.

Taken together, our recommendation is to screen models locally, and then to validate the actual model file, running as it would in production, on the intended computer. Finally, these results speak to which models are good candidates and what limits their deployment. They do not show that any of these models makes autonomous flight safe, or that it saves energy.

\FloatBarrier\end{document}